\documentclass[conference]{IEEEtran}
\IEEEoverridecommandlockouts

\usepackage{cite}
\usepackage{amsmath,amssymb,amsfonts}
\usepackage{algorithmic}
\usepackage{graphicx}
\usepackage{textcomp}
\usepackage{xcolor}
\usepackage{tabularx}
\usepackage{booktabs}

\def\BibTeX{{\rm B\kern-.05em{\sc i\kern-.025em b}\kern-.08em
    T\kern-.1667em\lower.7ex\hbox{E}\kern-.125emX}}
\begin{document}

\title{Textile Analysis for Recycling Automation using Transfer Learning \& Zero-Shot Foundation Models\\
{\footnotesize}
}

\author{
\IEEEauthorblockN{Yannis Spyridis}
\IEEEauthorblockA{
\textit{Department of Computer Science} \\
\textit{Kingston University}, London, UK \\
y.spyridis@kingston.ac.uk
}
\and
\IEEEauthorblockN{Vasileios Argyriou}
\IEEEauthorblockA{
\textit{Department of Networks and Digital Media} \\
\textit{Kingston University}, London, UK \\
vasileios.argyriou@kingston.ac.uk
}
}

\maketitle

\begin{abstract}
Automated sorting is crucial for improving the efficiency and scalability of textile recycling, but accurately identifying material composition and detecting contaminants from sensor data remains challenging. This paper investigates the use of standard RGB imagery, a cost-effective sensing modality, for key pre-processing tasks in an automated system. We present computer vision components designed for a conveyor belt setup to perform (a) classification of four common textile types and (b) segmentation of non-textile features such as buttons and zippers. For classification, several pre-trained architectures were evaluated using transfer learning and cross-validation, with EfficientNetB0 achieving the best performance on a held-out test set with 81.25\% accuracy. For feature segmentation, a zero-shot approach combining the Grounding DINO open-vocabulary detector with the Segment Anything Model (SAM) was employed, demonstrating excellent performance with a mIoU of 0.90 for the generated masks against ground truth. This study demonstrates the feasibility of using RGB images coupled with modern deep learning techniques, including transfer learning for classification and foundation models for zero-shot segmentation, to enable essential analysis steps for automated textile recycling pipelines.
\end{abstract}


\section{Introduction}
The global textile industry faces significant sustainability challenges, with vast quantities of post-consumer textiles ending up in landfills or incineration annually \cite{wojnowska2024growing}. This linear model consumes substantial resources, including water, energy, and raw materials, while contributing to pollution and greenhouse gas emissions \cite{leal2024reducing}. Transitioning towards a circular economy, where textiles are reused and recycled, is imperative for mitigating environmental impact and unlocking economic value from waste streams. Effective recycling requires accurate sorting of textiles based on material composition, a process that remains a major bottleneck. Automating this sorting process is crucial for increasing the efficiency, scalability, and economic viability of textile recycling initiatives worldwide \cite{alpert2021scalability}.

Current textile sorting is predominantly manual, a process that is labor-intensive, time-consuming, and often prone to inconsistencies and errors, especially when dealing with blended fabrics or complex garments \cite{ben2023opportunities,v04}. The speed of manual sorting limits the throughput of recycling facilities, while the associated labor costs can hinder profitability. Furthermore, manual sorting may not always reliably identify non-textile components like buttons and zippers, which must be removed prior to mechanical or chemical recycling processes to avoid contaminating the recycled material or damaging equipment \cite{behrendt2023contamination}. Therefore, there is a pressing need for automated systems capable of rapidly and accurately identifying textile materials and detecting problematic non-textile features.

This paper addresses these challenges by proposing a computer vision system designed for automated textile analysis within a recycling pipeline. The system utilises standard RGB images captured from a camera mounted above a conveyor belt, simulating a realistic industrial sorting environment. As textile items pause on the belt, images are acquired and processed by two specialised deep learning models. The first is a classification model trained to identify four common textile categories critical for recycling streams: Cotton, Polyester, Cotton-Polyester blends, and Viscose-Polyester blends. The second is a segmentation model designed to precisely locate non-textile features, specifically buttons and zippers, enabling their potential automated removal. Importantly, this work relies exclusively on real-world image data captured via standard sensors, reflecting practical operational conditions.

The main contributions of this paper are the following: Firstly, we demonstrate the feasibility of using cost-effective RGB imagery for the challenging tasks of classifying specific textile blends and segmenting small, potentially low-contrast features like buttons and zippers. Secondly, we present the design and implementation of tailored classification and segmentation models integrated into a workflow representative of an automated pre-processing stage for textile recycling. Thirdly, we validate the performance of these models using a dataset collected under realistic conditions, providing benchmarks for these specific tasks. Importantly, this work provides essential computer vision capabilities that can enable downstream automated processes, such as robotic sorting or targeted laser cutting for contaminant removal. 

The remainder of this paper is organised as follows: Section \ref{sec:related_work} reviews related literature in textile sorting and computer vision applications. Section \ref{sec:methodology} details our data acquisition, dataset characteristics, preprocessing steps, model architectures, and evaluation metrics. Section \ref{sec:results} presents the experimental results for both classification and segmentation tasks. Lastly, Section \ref{sec:conclusion} concludes the paper.

\section{Related Work}
\label{sec:related_work}

Automated analysis of textile materials is essential for advancing efficient sorting and recycling processes within the textile industry \cite{stanescu2021state}. Traditional manual sorting suffers from limitations in speed, cost, and consistency, motivating the development of instrumental techniques. Spectroscopic methods, particularly Near-Infrared (NIR) spectroscopy, combined with chemometrics and machine learning classifiers such as SVMs, have been widely investigated and proven effective for identifying textile fiber composition \cite{zhou2019textile, peets2017identification, sun2016classification, riba2020circular, du2022efficient, riba2022post}. Other approaches might include hyperspectral imaging and photometric techniques providing detailed surface information \cite{argyriou2010photometric}. While powerful, these methods often require specialised, potentially costly equipment and may face challenges related to sensor calibration and data fusion if multiple modalities are used.

In parallel, Computer Vision (CV) using standard RGB cameras offers a potentially lower-cost and more accessible alternative for analysing textiles. CV has been applied to various tasks in the textile industry, such as quality control and automated defect detection \cite{haleem2021computer, dlamini2022development, v01, v02, v03}. However, identifying textile material composition purely from RGB images is challenging due to subtle visual differences between materials and variations in texture, color, and lighting. Some studies have explored RGB-based classification for specific fabric colours \cite{9206901, zhou2022computer}. Convolutional Neural Networks (CNNs), including architectures like VGG \cite{simonyan2015deepconvolutionalnetworkslargescale} and EfficientNet \cite{tan2020efficientnetrethinkingmodelscaling}, have become the standard for image classification tasks, often leveraged effectively through transfer learning from models pre-trained on large datasets like ImageNet \cite{ridnik2021imagenet, gupta2022deep}. Applying these techniques effectively to distinguish common garment materials, including visually similar blends, remains an active area of investigation.

Beyond material identification, preparing textiles for recycling often requires the detection and localisation of non-textile contaminants such as buttons and zippers. This falls under the domain of object detection and semantic segmentation in computer vision. Established methods often rely on training CNN-based models like Faster R-CNN, YOLO, Mask R-CNN, or U-Net on large datasets with specific annotations for the objects of interest \cite{manakitsa2024review, chen2024review, vijayakumar2024yolo}. Recently, a new paradigm has emerged with Foundation Models, which are large models pre-trained on vast datasets that can be adapted to various downstream tasks with minimal or zero task-specific training data. For segmentation, the Segment Anything Model (SAM) provides powerful promptable segmentation capabilities \cite{kirillov2023segment}. SAM can generate masks based on various prompts, such as points, bounding boxes or previous masks. Recently, models like Grounding DINO \cite{liu2024grounding} have been introduced that can accept text prompts to perform open-vocabulary object detection, enabling the localisation of objects based on arbitrary descriptions without requiring explicit training on those categories. These text-guided detectors can provide bounding boxes that serve as prompts for segmentation models such as SAM. Applying such zero-shot detection and segmentation pipelines, especially for small objects like clothing fasteners in industrial settings can offer a novel alternative to traditional supervised approaches.

Developing and rigorously evaluating models for these tasks, whether through supervised training or applying pre-trained systems, necessitates relevant datasets. A significant challenge is the scarcity of large, publicly available datasets specifically curated for RGB-based textile material identification and contaminant detection relevant to recycling. While datasets exist for fabric defect detection \cite{li2021fabric, bergmann2021mvtec}, they do not address material type classification or feature segmentation needed for sorting pre-processing. This lack of public data hinders the development and benchmarking of robust algorithms.

\section{Methodology}
\label{sec:methodology}
This section details the experimental methodology employed in this study. We describe the process for acquiring the image data, the characteristics of the resulting dataset, the preprocessing steps applied, the architectures of the classification and segmentation models, and the training procedures. Performance evaluation was conducted using a 5-fold cross-validation strategy to ensure robust assessment despite the dataset size, and standard metrics were used to quantify model performance.

\subsection{Data Acquisition}

The image data used in this work was collected to simulate conditions representative of an automated textile sorting facility utilising a conveyor belt system. Textile items were placed flat on the conveyor surface. Image capture was triggered when the belt was stationary, ensuring motion blur was minimised. A modern smartphone camera was used as the acquisition device, featuring a 1/1.33-inch sensor, an f/1.8 aperture, and configured to capture images at a resolution of 12 megapixels. This choice reflects the potential for using accessible components in practical systems. To maintain consistency across the dataset, all images were captured under similar ambient indoor lighting conditions. Crucially, no post-processing or digital enhancements were applied to the raw images after capture, preserving the data's fidelity to the real-world sensor output.

\subsection{Dataset}

The acquired images form the basis of the dataset used for developing and evaluating the computer vision models, comprising exclusively real-world RGB images obtained as described above. The total dataset consists of 80 unique images. For the textile classification task, the dataset supported identification across four key material types relevant to recycling: Cotton, Polyester, Cotton-Polyester blend, and Viscose-Polyester blend. There were 20 distinct images available for each of these four classes, yielding a balanced dataset for training and validation.

gmentation task, images containing buttons and zippers were used. These features were manually annotated within the images to create ground truth masks, enabling evaluation of segmentation performance for pixel-level localisation. All images in the dataset were manually annotated for these features, creating pixel-level ground truth masks to serve as reference for evaluating the segmentation results.

Given the limited size of the overall dataset, a 5-fold cross-validation strategy was employed for the classification model training and validation to maximise the utility of the available data and provide a more reliable estimate of generalisation performance. The 80 images were randomly partitioned into 5 mutually exclusive folds, each containing 16 images, with class balance maintained within folds. For each fold iteration, 80\% of the data was used for training and 20\% for validation, rotating through all folds. This approach ensured that each image was used for validation exactly once across the 5 folds.

For the final testing phase of the classification models, a separate test set was constructed with 6 images each from the Cotton, Polyester, and Cotton-Polyester classes. For the Viscose-Polyester class, we included additional samples - 14 total in the test set. This resulted in a final test set of 32 images that remained completely separate from the training and validation process, ensuring an unbiased evaluation of model generalisation. The final performance metrics for the classification task reported in Section \ref{sec:results} represent the results on this dedicated test set, complemented by the insights gained from the 5-fold cross-validation during model development.

\subsection{Data Preprocessing}
\label{sec:preprocessing}
Input images were systematically preprocessed to ensure suitability for the different neural network architectures employed in the classification task. Raw images were first loaded and then spatial normalisation was performed to match the specific input dimensions required by each model, as detailed in Table \ref{tab:model_input_dims}. This involved extracting a central crop from the image, sized according to the minimum of the original dimensions and the target dimensions, followed by resizing this crop to the exact target resolution using bicubic interpolation.

\begin{table}[b!]
\centering
\caption{Classification model architectures and corresponding input dimensions}
\begin{tabularx}{\linewidth}{@{}X X@{}}
\toprule
\textbf{Model Architecture} & \textbf{Input Dimensions (pixels)} \\
\midrule
VGG16               & 224 $\times$ 224 \\
VGG19               & 224 $\times$ 224 \\
EfficientNetB0      & 224 $\times$ 224 \\
EfficientNetV2-S    & 384 $\times$ 384 \\
EfficientNetV2-M    & 480 $\times$ 480 \\
\bottomrule
\end{tabularx}
\label{tab:model_input_dims}
\end{table}

After spatial resizing and color space conversion, pixel values were linearly scaled to the range [-1, 1]. This normalisation step ensures that input data adheres to the expected value range for the models utilised.

To enhance model generalisation and mitigate overfitting, particularly given the dataset size, data augmentation was applied exclusively during the training phase of the classification models. A set of random transformations were applied dynamically to the training images within each batch. The specific augmentation techniques and their corresponding parameters are outlined in Table \ref{tab:augmentation_params}.

\begin{table}[t!]
\centering
\caption{Applied data augmentation techniques during training}
\begin{tabularx}{\linewidth}{@{}X X@{}}
\toprule
\textbf{Augmentation technique} & \textbf{Parameter range} \\
\midrule
Random Rotation         & Up to 90 degrees \\
Random Width Shift      & Up to 30\% of image width \\
Random Height Shift     & Up to 30\% of image height \\
Random Zoom             & Up to 30\% (zoom in or out) \\
Random Horizontal Flip  & Applied with 50\% probability \\
\bottomrule
\end{tabularx}
\label{tab:augmentation_params}
\end{table}

\subsection{Textile Classification Models and Training}
For the textile classification task, we evaluated the performance of five deep convolutional neural network architectures, leveraging transfer learning from models pre-trained on the ImageNet dataset. The selected architectures were VGG16, VGG19, EfficientNetB0, EfficientNetV2-S, and EfficientNetV2-M. These models offer a range of computational complexities and have demonstrated strong performance on various vision tasks. The specific input image dimensions required for each model were used during preprocessing as detailed in \ref{sec:preprocessing}.

For each architecture, the convolutional base was initialised with its ImageNet pre-trained weights. The original classification layer was removed and replaced with a new head suitable for our 4-class textile identification problem, consisting of a Global Average Pooling 2D layer followed by a Dense layer with Softmax activation.

A two-phase training strategy was employed within the 5-fold cross-validation framework to effectively adapt the pre-trained models to our specific dataset:

\textbf{Phase 1}: Feature extraction: Initially, only the weights of the newly added classification head were trained, while the weights of the pre-trained convolutional base were kept frozen. This phase aimed to adapt the classifier to the features extracted by the base model. This phase ran for a maximum of 15 epochs using the Adam optimiser with a learning rate of \( 1 \text{e} - 3 \).

\textbf{Phase 2}: Fine-tuning: Following the initial phase, a portion of the pre-trained base model layers were unfrozen to allow for fine-tuning along with the classification head, enabling the model to adapt features more closely to the textile dataset. Specifically, the top 30 layers were unfrozen for the EfficientNet models  while for the VGG models, the top 4 layers (VGG16) and top 6 layers (VGG19) were unfrozen respectively.  Training continued for a maximum of 50 additional epochs using the Adam optimiser with a significantly reduced learning rate of \( 1 \text{e} - 5 \). This fine-tuning phase allows the model to adapt the extracted features more closely to the specifics of the textile dataset.
Throughout both training phases, the Categorical Cross-Entropy loss function was minimised. Training was performed using a batch size of 4. Several callbacks were employed to manage the training process within each fold:

To optimise training and prevent overfitting, several callbacks were employed:

\begin{itemize}
    \item \textbf{Early Stopping}: Monitored the validation loss with a patience of 15 epochs. Training was halted if no improvement was observed, and the weights from the epoch with the lowest validation loss were restored.
    \item \textbf{Learning Rate Reduction on Plateau}: Monitored the validation loss and reduced the learning rate by a factor of $0.2$ after $10$ consecutive epochs with no improvement, with a minimum threshold of $1\times10^{-6}$.
    \item \textbf{Model Checkpointing}: Saved the weights corresponding to the highest validation accuracy across both training phases for each cross-validation fold.
\end{itemize}

All reported results in Section~\ref{sec:results} are based on the best-performing fold for each model. The key hyperparameters used for training the classification models are summarised in Table \ref{tab:training_config}.

\begin{table}[t!]
\centering
\caption{Model training configuration and hyperparameter settings}
\begin{tabularx}{\linewidth}{@{}X X@{}}
\toprule
\textbf{Hyperparameter} & \textbf{Value / Setting} \\
\midrule
Pre-trained Weights         & ImageNet \\
Classification Head         & Global Average Pooling 2D + Dense (4 units, Softmax) \\
Training Phases             & 1: Feature Extraction (head only), 2: Fine-tuning (head + top 30 base layers (4-6 for VGG) \\
Optimiser                   & Adam \\
Learning Rate (Phase 1)     & \( 1 \text{e} - 3 \) \\
Learning Rate (Phase 2)     & \( 1 \text{e} - 5 \) \\
Loss Function               & Categorical Cross-Entropy \\
Batch Size                  & 4 \\
Max Epochs (Phase 1)        & 15 \\
Max Epochs (Phase 2)        & 50 \\
Early Stopping              & Monitor: \texttt{val\_loss}, Patience: 15, Restore Best Weights: True \\
Reduce LR on Plateau        & Monitor: \texttt{val\_loss}, Patience: 10, Factor: 0.2, Min LR: \( 1 \text{e} - 6 \) \\
Model Checkpoint            & Monitor: \texttt{val\_accuracy}, Save Best Only: True \\
Evaluation Framework        & 5-Fold Cross-Validation \\
\bottomrule
\end{tabularx}
\label{tab:training_config}
\end{table}

\subsection{Feature Segmentation using Grounding DINO and SAM}
For the task of identifying and locating non-textile features within the garment images, a methodology utilising state-of-the-art pre-trained foundation models was employed, circumventing the need for training a dedicated segmentation network on this specific dataset. This approach utilised a combination of Grounding DINO \cite{liu2024grounding} for text-prompted open-vocabulary object detection and the SAM model \cite{kirillov2023segment} for high-fidelity prompt-based segmentation.

The operational pipeline proceeded as follows: Each input image was processed alongside specific textual prompts corresponding to the features of interest – primarily "button" and "zipper". The Grounding DINO model interpreted these prompts to perform open-set detection, identifying candidate regions within the image likely corresponding to the textual descriptions. The output from this stage consisted of bounding boxes delineating the detected instances of buttons and zippers. Standard confidence thresholds associated with the detector could be applied to filter detections, though default parameters were generally sufficient for this application.

Subsequently, the bounding boxes generated by Grounding DINO served as input prompts to the SAM model. SAM, conditioned on the original image and these spatial bounding box prompts, then generated precise pixel-level segmentation masks and corresponding object contours for the indicated objects. This output precisely demarcates the boundaries and pixel regions belonging to the detected buttons or zippers.

This synergistic pipeline utilises the zero-shot capabilities of Grounding DINO to locate objects based on language descriptions and the powerful segmentation abilities of SAM guided by spatial cues, thereby enabling the segmentation of target features without task-specific training. The primary advantage of this approach lies in its ability to leverage large-scale pre-training, potentially reducing the need for extensive annotated data typically required for supervised segmentation model training, although annotated data remains essential for evaluation.

The performance of this feature segmentation pipeline was quantitatively evaluated by comparing the predicted segmentation masks (or the area implied by the contours) against the manually created pixel-level ground truth annotations (described in Section 3.2). The evaluation, using standard segmentation metrics detailed in Section 3.6, was conducted across all 80 images in the dataset containing these annotated features.

\subsection{Evaluation Metrics}
To quantitatively assess the performance of the developed textile classification models and the feature segmentation approach, standard and well-established evaluation metrics were employed. Separate metrics appropriate for multi-class classification and pixel-level segmentation tasks were used.

\subsubsection{Classification Metrics} 

The performance of the textile classification models was evaluated using metrics derived from the aggregation of results across the 5 folds of the cross-validation procedure. These metrics rely on the counts of True Positives (TP), True Negatives (TN), False Positives (FP), and False Negatives (FN), considered on a per-class basis.

\textbf{Accuracy}: Represents the overall fraction of correctly classified instances across all classes.
\begin{equation}
\text{Accuracy} = \frac{\text{Number of correct predictions}}{\text{Total number of predictions}}
\end{equation}

\textbf{Precision}: Measures the proportion of instances predicted as belonging to class $i$ that actually belong to class $i$. It reflects the exactness of the prediction.
\begin{equation}
\text{Precision}_i = \frac{TP_i}{TP_i + FP_i}
\end{equation}

\textbf{Recall}: Measures the proportion of actual instances of class $i$ that were correctly identified by the model. It reflects the completeness or coverage of the class.
\begin{equation}
\text{Recall}_i = \frac{TP_i}{TP_i + FN_i}
\end{equation}

\textbf{F1-Score}: The harmonic mean of Precision and Recall, providing a single score that balances both metrics. It is particularly useful when class distributions are uneven.
\begin{equation}
\text{F1}_i = 2 \times \frac{\text{Precision}_i \times \text{Recall}_i}{\text{Precision}_i + \text{Recall}_i}
\end{equation}

\subsubsection{Segmentation Metrics}
The performance of the feature segmentation pipeline was evaluated by comparing the predicted pixel-level masks for buttons and zippers against the manual ground truth annotations across all relevant images in the dataset. The evaluation of the segmentation approach operates at the pixel level, and relies on metrics that measure how well the predicted segmentation masks match the ground truth masks on a per-pixel basis. The following standard metrics were employed:

\textbf{Intersection over Union (IoU)}: IoU measures the overlap between the predicted and ground truth segmentation masks. It is defined as the ratio of the intersection to the union of the predicted and actual pixel sets:
\begin{equation}
\text{IoU}_i = \frac{TP_i}{TP_i + FP_i + FN_i}
\end{equation}

\section{Results and Experiments}
\label{sec:results}

This section details the experimental results obtained for both the textile classification and feature segmentation tasks. We present quantitative performance metrics based on the best fold of the 5-fold cross-validation for classification, followed by the results of the feature segmentation analysis.

\subsection{Classification Performance}
Following the 5-fold cross-validation training and selection process described in Section \ref{sec:methodology}, the best performing fold's checkpoint for each of the five architectures was evaluated on unseen test data. The overall performance of these models on the test set is summarised in Table \ref{tab:classification_results}, comparing their accuracy and weighted average F1-scores.

\begin{table}[t!]
\centering
\caption{Overall classification performance. Results are reported for the best-performing fold (checkpoint) of each architecture.}
\label{tab:classification_results}
\begin{tabularx}{\linewidth}{@{}>{\hsize=1.25\hsize}X >{\hsize=0.75\hsize}X >{\hsize=1.00\hsize}X@{}}
\toprule
\textbf{Model Architecture} & \textbf{Accuracy} & \textbf{Weighted F1-Score} \\
\midrule
VGG16 (Fold 5)            & 0.4688            & 0.4646 \\
VGG19 (Fold 4)            & 0.4375            & 0.4483 \\
EfficientNetB0 (Fold 3)   & \textbf{0.8125}   & \textbf{0.8012} \\
EfficientNetV2-S (Fold 3) & \textbf{0.8125}   & 0.7927 \\
EfficientNetV2-M (Fold 2) & 0.6562            & 0.6062 \\
\bottomrule
\end{tabularx}
\end{table}

\begin{table}[t!]
\centering
\caption{Per-class F1-scores on the test data. Results are reported for the best-performing fold of each architecture.}
\label{tab:per_class_f1}
\begin{tabularx}{\linewidth}{@{}>{\hsize=0.7\hsize}X>{\hsize=0.375\hsize}X>{\hsize=0.375\hsize}X>{\hsize=0.375\hsize}X>{\hsize=0.375\hsize}X@{}}
\toprule
\textbf{Model} & \textbf{Cotton} & \textbf{Polyester} & \textbf{Cotton-Polyester} & \textbf{Viscose-Polyester} \\
\midrule
VGG16            & 0.4545 & 0.2857 & 0.1818 & 0.6667 \\
VGG19            & 0.4444 & 0.4615 & 0.0000 & 0.6364 \\
EfficientNetB0   & \textbf{0.8571} & 0.7692 & 0.4000 & \textbf{0.9630} \\
EfficientNetV2-S & 0.7500 & \textbf{0.8000} & \textbf{0.5000} & 0.9333 \\
EfficientNetV2-M & 0.6000 & \textbf{0.8000} & 0.0000 & 0.7857 \\
\bottomrule
\end{tabularx}
\end{table}

The table clearly shows that the EfficientNet models significantly outperformed the VGG architectures on the test data. EfficientNetB0 and EfficientNetV2S achieved the highest accuracy (81.25\%), with EfficientNetB0 also obtaining the highest weighted F1-score (0.8012), suggesting it provided the best balance of precision and recall across the classes in the test set. EfficientNetV2M performed less well, while both VGG models failed to achieve even 50\% accuracy or F1-score. To provide a more granular view of performance across the different textile types, Table \ref{tab:per_class_f1} presents the per-class F1-scores for each model on the test set.

The per-class F1-scores in Table \ref{tab:per_class_f1} reveal specific strengths and weaknesses. The top-performing EfficientNetB0 model excelled particularly on Cotton (0.8571) and the most frequent class, Viscose-Polyester (0.9630), but showed moderate difficulty with Cotton-Polyester (0.4000). EfficientNetV2S, while having the same overall accuracy, achieved better F1-scores for Polyester (0.8000) and Cotton-Polyester (0.5000) than EfficientNetB0, but lagged slightly on Cotton and Viscose-Polyester. Notably, both EfficientNetV2M and VGG19 completely failed to correctly identify any Cotton-Polyester samples in the test set, highlighting this blend as a significant challenge. The VGG models generally showed poor F1-scores across most classes.

\begin{figure}[b!]
    \centering
    \includegraphics[width=0.7\linewidth]{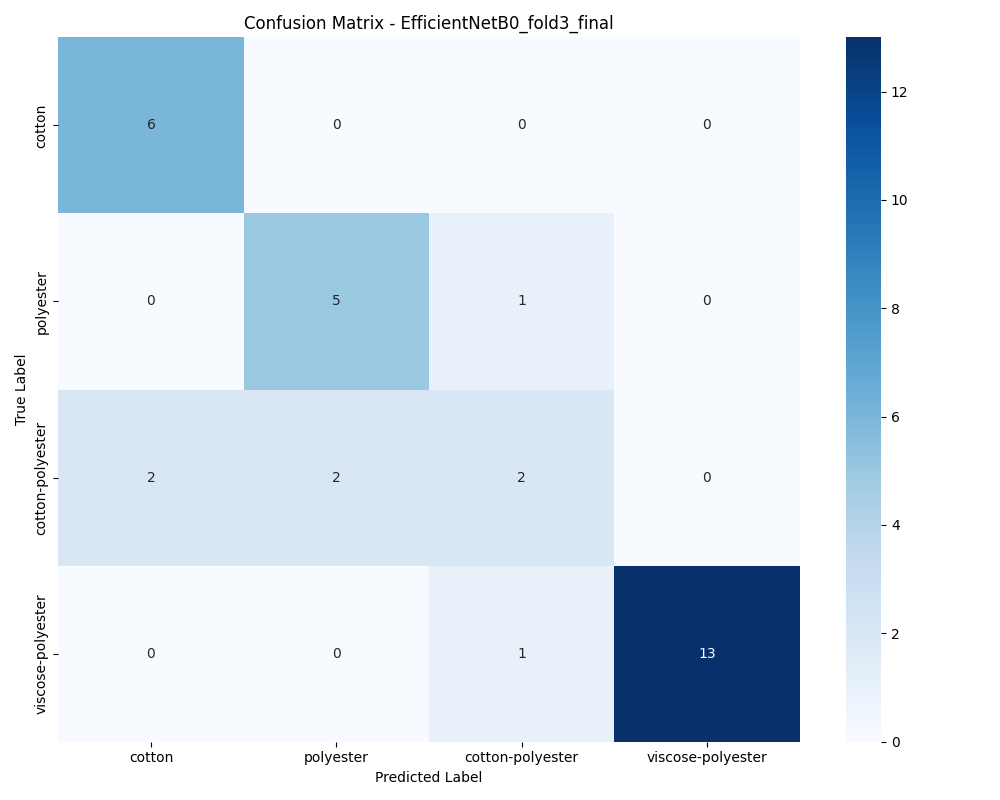}
    \caption{Confusion matrix for EfficientNetB0 (Fold 3)}
    \label{fig:effb0_cm}
\end{figure}

The confusion matrix for the EfficientNetB0 model is depicted in Figure \ref{fig:effb0_cm} and visually summarises the classification performance of the best model on the test set. It confirms excellent results for Cotton with all samples correctly classified and Viscose-Polyester with 13/14 correctly classified, with the latter also demonstrating perfect precision as no other class was misclassified as Viscose-Polyester. The primary difficulty occurred with the Cotton-Polyester blend, where only 2 of its 6 instances were correctly identified, with 2 being misclassified as Cotton and 2 as Polyester. A minor confusion also occurred where 1 Polyester instance was misclassified as Cotton-Polyester. This highlights the challenge in reliably differentiating the Cotton-Polyester blend from pure Cotton and Polyester in the dataset.

\subsection{Classification Model Training Analysis}

\begin{figure}[b!]
    \centering
    \includegraphics[width=0.8\linewidth]{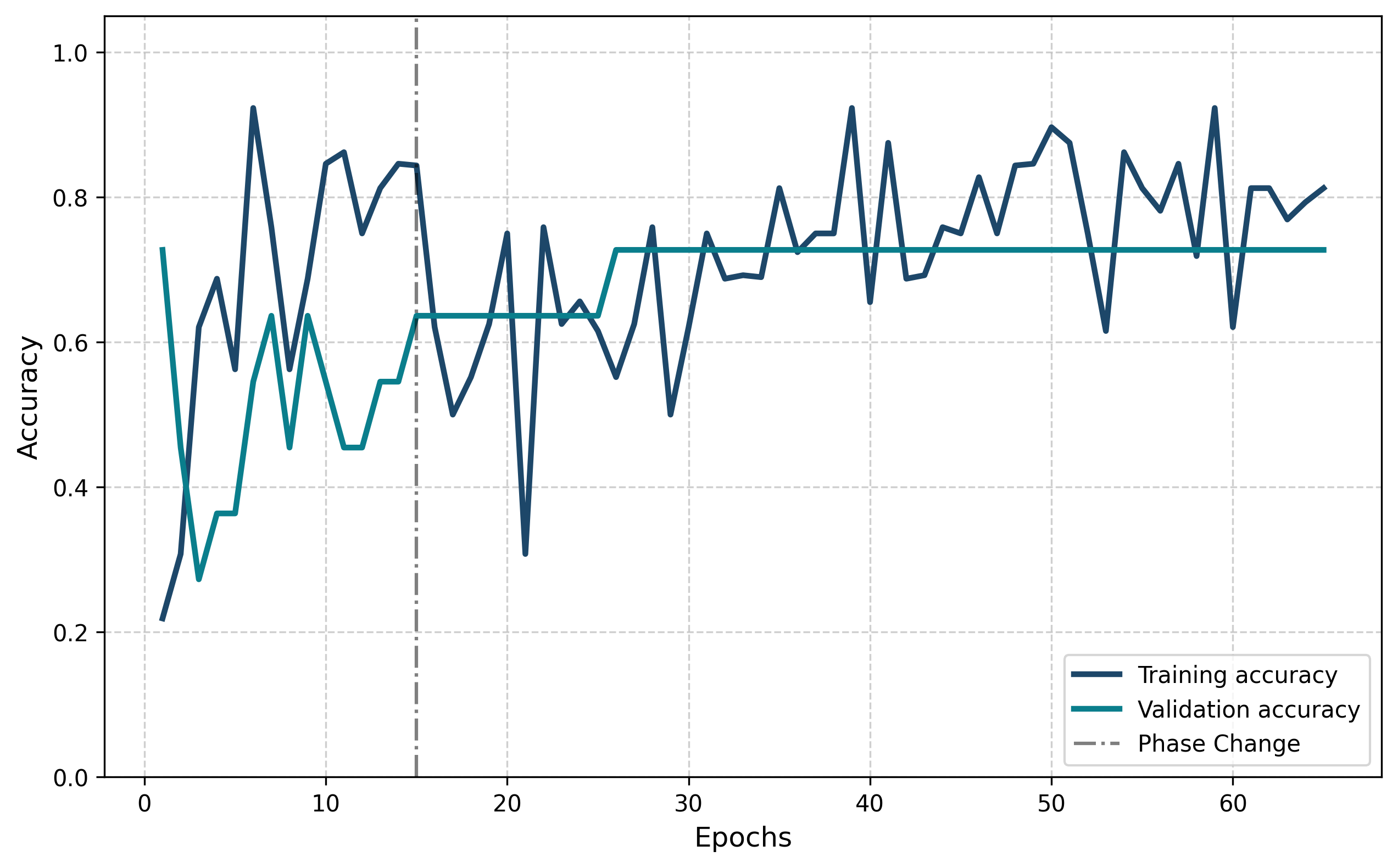}
    \caption{Training and validation accuracy for EfficientNetB0 (Fold 3)}
    \label{fig:train_val_accuracy}
\end{figure}

\begin{figure}[b!]
    \centering
    \includegraphics[width=0.8\linewidth]{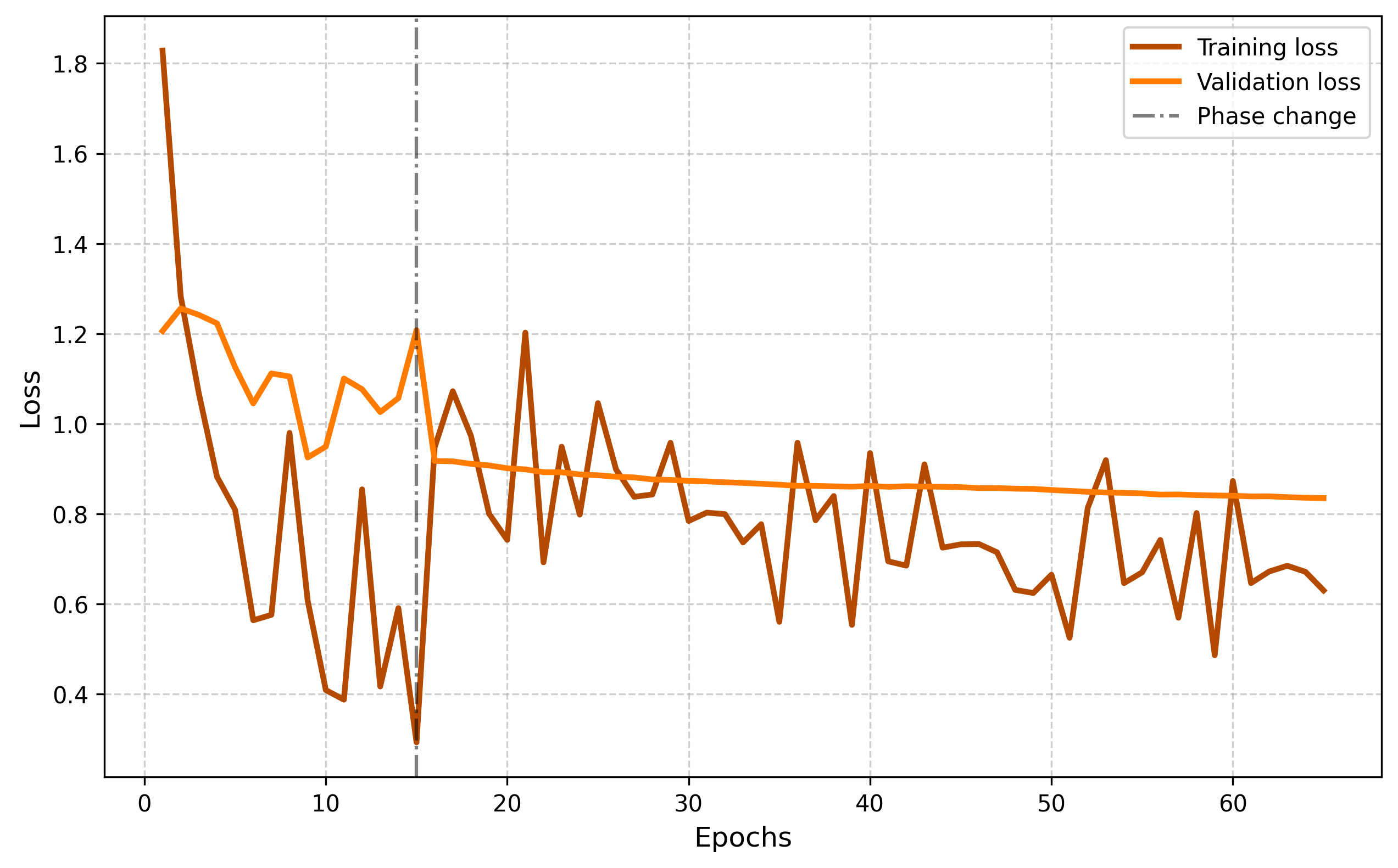}
    \caption{Training and validation loss for EfficientNetB0 (Fold 3)}
    \label{fig:train_val_loss}
\end{figure}

\begin{figure*}[t!]
    \centering
    \includegraphics[width=0.8\linewidth]{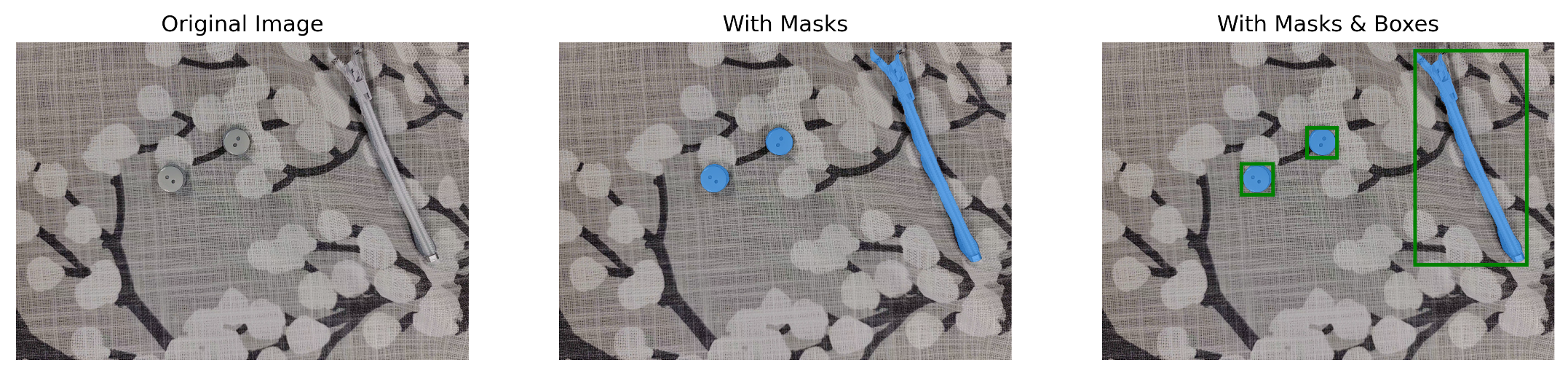}
    \caption{Example segmentation results}
    \label{fig:seg_images}
\end{figure*}

To provide insight into the learning process of the best-performing model identified during test set evaluation, the training and validation accuracy and loss curves for the EfficientNetB0 Fold 3 checkpoint are presented in Figures \ref{fig:train_val_accuracy} and \ref{fig:train_val_loss} respectively. These curves illustrate the model's behaviour throughout the two-phase training, within that specific cross-validation fold. The figures illustrate the model's successful convergence during training, indicated by the upward trend in accuracy and downward trend in loss. While exhibiting an expected gap between training and validation performance for this dataset size, the validation loss stabilised, suggesting appropriate regularisation via early stopping.

While the checkpoint from Fold 3 yielded the best performance on the final test set, it is worth noting that a different Fold, namely Fold 4 achieved higher peak validation accuracies during the cross-validation process. This illustrates the expected variance between validation performance observed during training and final generalisation performance on the separate test set, particularly in our small dataset.

\subsection{Feature Segmentation Performance}

The performance of the feature segmentation pipeline, utilising Grounding DINO for text-prompted detection and SAM for mask generation, was evaluated on the task of identifying buttons and zippers across the dataset. Quantitative evaluation focused on the quality of the generated bounding boxes and segmentation masks compared to manual annotations. Table \ref{tab:segmentation_results} outlines the results.

\begin{table}[t!]
\centering
\caption{Performance metrics for the feature segmentation pipeline.}
\label{tab:segmentation_results}
\begin{tabularx}{\linewidth}{@{}>{\hsize=0.875\hsize}X
                              >{\hsize=1.25\hsize}X
                              >{\hsize=0.875\hsize}X@{}}
\toprule
\textbf{Metric} & \textbf{Bounding Box Prediction} & \textbf{Mask Prediction} \\
\midrule
Precision       & 1.00 & 1.00 \\
Recall          & 1.00 & 1.00 \\
F1-Score        & 1.00 & 1.00 \\
Mean IoU        & 0.92 & 0.90 \\
\bottomrule
\end{tabularx}
\end{table}

The results indicated highly effective performance for this approach on the collected dataset. The mean Intersection over Union, measuring the overlap between predicted and ground truth regions, was calculated. For the final segmentation masks, the pipeline achieved a Mean IoU of 0.90. The intermediate bounding box predictions also showed strong localisation with a Mean IoU of 0.92. Furthermore, object-level detection metrics, including Precision, Recall, and F1-score, were reported as 1.00 for both mask and bounding box predictions, suggesting that essentially all target features were correctly detected with no false positives.

Qualitative examples illustrating the pipeline's output are shown in Figure \ref{fig:seg_images}. These examples display the original textile image alongside the generated segmentation masks and bounding boxes for detected buttons and zippers, visually confirming the precise localisation achieved by the method.

\section{Conclusion}
\label{sec:conclusion}

This paper addressed the need for automated analysis of textile materials from standard RGB images to support efficient textile recycling. We developed and evaluated computer vision components for two key tasks: classifying common textile materials, including challenging blends, and segmenting non-textile contaminants designated for removal. For classification, several pre-trained architectures were investigated using transfer learning and a 5-fold cross-validation approach on a custom dataset, with EfficientNetB0 demonstrating the best performance on an unseen test set, achieving 81.25\% accuracy. For feature segmentation, a zero-shot methodology combining the Grounding DINO open-vocabulary detector \& the Segment Anything Model was employed, yielding excellent results with a Mean IoU of 0.9 for the generated masks.

The contributions of this work lie in demonstrating the feasibility of using readily available RGB camera data and modern deep learning techniques for these specific, practical tasks within the textile recycling domain. We established performance benchmarks for classifying common materials, including blends, and successfully applied a state-of-the-art foundation model pipeline for accurate contaminant segmentation without task-specific training. These developed vision capabilities represent essential enabling technologies for automating critical pre-processing steps, such as material sorting and contaminant removal, thereby paving the way for more scalable and cost-effective automated textile recycling systems.

\section*{Acknowledgement}
This work was funded by UK Research and Innovation (UKRI) under the UK government’s Innovate UK Smart Grants [grant number 10111591 (ReFibres)].

\bibliography{ref}

\begin{thebibliography}{10}
\providecommand{\url}[1]{#1}
\csname url@samestyle\endcsname
\providecommand{\newblock}{\relax}
\providecommand{\bibinfo}[2]{#2}
\providecommand{\BIBentrySTDinterwordspacing}{\spaceskip=0pt\relax}
\providecommand{\BIBentryALTinterwordstretchfactor}{4}
\providecommand{\BIBentryALTinterwordspacing}{\spaceskip=\fontdimen2\font plus
\BIBentryALTinterwordstretchfactor\fontdimen3\font minus \fontdimen4\font\relax}
\providecommand{\BIBforeignlanguage}[2]{{%
\expandafter\ifx\csname l@#1\endcsname\relax
\typeout{** WARNING: IEEEtran.bst: No hyphenation pattern has been}%
\typeout{** loaded for the language `#1'. Using the pattern for}%
\typeout{** the default language instead.}%
\else
\language=\csname l@#1\endcsname
\fi
#2}}
\providecommand{\BIBdecl}{\relax}
\BIBdecl

\bibitem{wojnowska2024growing}
I.~Wojnowska-Bary{\l}a, K.~Bernat, M.~Zaborowska, and D.~Kulikowska, ``The growing problem of textile waste generation—the current state of textile waste management,'' \emph{Energies}, vol.~17, no.~7, p. 1528, 2024.

\bibitem{leal2024reducing}
W.~Leal~Filho, M.~A.~P. Dinis, O.~Liakh, A.~Pa{\c{c}}o, K.~Dennis, F.~Shollo, and H.~Sidsaph, ``Reducing the carbon footprint of the textile sector: an overview of impacts and solutions,'' \emph{Textile Research Journal}, vol.~94, no. 15-16, pp. 1798--1814, 2024.

\bibitem{alpert2021scalability}
C.~Alpert, M.~Turkowski, and T.~Tasneem, ``Scalability solutions for automated textile sorting: a case study on how dynamic capabilities can overcome scalability challenges,'' 2021.

\bibitem{ben2023opportunities}
R.~Ben~Amor, K.~T.~W. Ng, T.~T. Sithi, and T.~S. Mahmud, ``Opportunities and challenges for the sorting of post-consumer textile waste,'' in \emph{Canadian Society of Civil Engineering Annual Conference}.\hskip 1em plus 0.5em minus 0.4em\relax Springer, 2023, pp. 89--99.

\bibitem{v04}
J.~Clark, G.~Johnson, O.~Duran, and V.~Argyriou, ``Fabric composition classification using hyper-spectral imaging,'' in \emph{2023 19th International Conference on Distributed Computing in Smart Systems and the Internet of Things (DCOSS-IoT)}, 2023, pp. 347--353.

\bibitem{behrendt2023contamination}
T.~Behrendt and E.~Eppinger, ``Contamination threshold values for textile recycling,'' in \emph{ITC-ICEE}.\hskip 1em plus 0.5em minus 0.4em\relax Springer, 2023, pp. 468--479.

\bibitem{stanescu2021state}
M.~D. Stanescu, ``State of the art of post-consumer textile waste upcycling to reach the zero waste milestone,'' \emph{Environmental Science and Pollution Research}, vol.~28, no.~12, pp. 14\,253--14\,270, 2021.

\bibitem{zhou2019textile}
J.~Zhou, L.~Yu, Q.~Ding, and R.~Wang, ``Textile fiber identification using near-infrared spectroscopy and pattern recognition,'' \emph{Autex Research Journal}, vol.~19, no.~2, pp. 201--209, 2019.

\bibitem{peets2017identification}
P.~Peets, I.~Leito, J.~Pelt, and S.~Vahur, ``Identification and classification of textile fibres using atr-ft-ir spectroscopy with chemometric methods,'' \emph{Spectrochimica Acta Part A: Molecular and Biomolecular Spectroscopy}, vol. 173, pp. 175--181, 2017.

\bibitem{sun2016classification}
X.~Sun, M.~Zhou, and Y.~Sun, ``Classification of textile fabrics by use of spectroscopy-based pattern recognition methods,'' \emph{Spectroscopy Letters}, vol.~49, no.~2, pp. 96--102, 2016.

\bibitem{riba2020circular}
J.-R. Riba, R.~Cantero, T.~Canals, and R.~Puig, ``Circular economy of post-consumer textile waste: Classification through infrared spectroscopy,'' \emph{Journal of Cleaner Production}, vol. 272, p. 123011, 2020.

\bibitem{du2022efficient}
W.~Du, J.~Zheng, W.~Li, Z.~Liu, H.~Wang, and X.~Han, ``Efficient recognition and automatic sorting technology of waste textiles based on online near infrared spectroscopy and convolutional neural network,'' \emph{Resources, Conservation and Recycling}, vol. 180, p. 106157, 2022.

\bibitem{riba2022post}
J.-R. Riba, R.~Cantero, P.~Riba-Mosoll, and R.~Puig, ``Post-consumer textile waste classification through near-infrared spectroscopy, using an advanced deep learning approach,'' \emph{Polymers}, vol.~14, no.~12, p. 2475, 2022.

\bibitem{argyriou2010photometric}
V.~Argyriou, M.~Petrou, and S.~Barsky, ``Photometric stereo with an arbitrary number of illuminants,'' \emph{Computer Vision and Image Understanding}, vol. 114, no.~8, pp. 887--900, 2010.

\bibitem{haleem2021computer}
N.~Haleem, M.~Bustreo, and A.~Del~Bue, ``A computer vision based online quality control system for textile yarns,'' \emph{Computers in Industry}, vol. 133, p. 103550, 2021.

\bibitem{dlamini2022development}
S.~Dlamini, C.-Y. Kao, S.-L. Su, and C.-F. Jeffrey~Kuo, ``Development of a real-time machine vision system for functional textile fabric defect detection using a deep yolov4 model,'' \emph{Textile Research Journal}, vol.~92, no. 5-6, pp. 675--690, 2022.

\bibitem{v01}
V.~Li, B.~Villarini, J.-C. Nebel, and A.~Vasileios, ``A modular deep learning framework for scene understanding in augmented reality applications,'' in \emph{2023 IEEE IAICT}.\hskip 1em plus 0.5em minus 0.4em\relax IEEE, 2023, pp. 45--51.

\bibitem{v02}
V.~Li, G.~Tsoumplekas, I.~Siniosoglou, V.~Argyriou, A.~Lytos, E.~Fountoukidis, and P.~Sarigiannidis, ``A closer look at data augmentation strategies for finetuning-based low/few-shot object detection,'' in \emph{2024 IEEE 14th International Symposium on Industrial Embedded Systems (SIES)}.\hskip 1em plus 0.5em minus 0.4em\relax IEEE, 2024, pp. 156--163.

\bibitem{v03}
V.~Li, I.~Siniosoglou, T.~Karamitsou, A.~Lytos, I.~D. Moscholios, S.~K. Goudos, J.~S. Banerjee, P.~Sarigiannidis, and V.~Argyriou, ``Enhancing 3d object detection in autonomous vehicles based on synthetic virtual environment analysis,'' \emph{IVC}, vol. 154, p. 105385, 2025.

\bibitem{9206901}
A.~C. da~Silva~BarrosM, E.~Firmeza~Ohata, S.~P.~P. da~Silva, J.~Silva~Almeida, and P.~P. Rebouças~Filho, ``An innovative approach of textile fabrics identification from mobile images using computer vision based on deep transfer learning,'' in \emph{2020 International Joint Conference on Neural Networks (IJCNN)}, 2020, pp. 1--8.

\bibitem{zhou2022computer}
J.~Zhou, X.~Zou, and W.~K. Wong, ``Computer vision-based color sorting for waste textile recycling,'' \emph{International Journal of Clothing Science and Technology}, vol.~34, no.~1, pp. 29--40, 2022.

\bibitem{simonyan2015deepconvolutionalnetworkslargescale}
\BIBentryALTinterwordspacing
K.~Simonyan and A.~Zisserman, ``Very deep convolutional networks for large-scale image recognition,'' 2015. [Online]. Available: \url{https://arxiv.org/abs/1409.1556}
\BIBentrySTDinterwordspacing

\bibitem{tan2020efficientnetrethinkingmodelscaling}
\BIBentryALTinterwordspacing
M.~Tan and Q.~V. Le, ``Efficientnet: Rethinking model scaling for convolutional neural networks,'' 2020. [Online]. Available: \url{https://arxiv.org/abs/1905.11946}
\BIBentrySTDinterwordspacing

\bibitem{ridnik2021imagenet}
T.~Ridnik, E.~Ben-Baruch, A.~Noy, and L.~Zelnik-Manor, ``Imagenet-21k pretraining for the masses,'' \emph{arXiv preprint arXiv:2104.10972}, 2021.

\bibitem{gupta2022deep}
J.~Gupta, S.~Pathak, and G.~Kumar, ``Deep learning (cnn) and transfer learning: a review,'' in \emph{Journal of Physics: Conference Series}, vol. 2273, no.~1.\hskip 1em plus 0.5em minus 0.4em\relax IOP Publishing, 2022, p. 012029.

\bibitem{manakitsa2024review}
N.~Manakitsa, G.~S. Maraslidis, L.~Moysis, and G.~F. Fragulis, ``A review of machine learning and deep learning for object detection, semantic segmentation, and human action recognition in machine and robotic vision,'' \emph{Technologies}, vol.~12, no.~2, p.~15, 2024.

\bibitem{chen2024review}
W.~Chen, J.~Luo, F.~Zhang, and Z.~Tian, ``A review of object detection: Datasets, performance evaluation, architecture, applications and current trends,'' \emph{Multimedia Tools and Applications}, vol.~83, no.~24, pp. 65\,603--65\,661, 2024.

\bibitem{vijayakumar2024yolo}
A.~Vijayakumar and S.~Vairavasundaram, ``Yolo-based object detection models: A review and its applications,'' \emph{Multimedia Tools and Applications}, vol.~83, no.~35, pp. 83\,535--83\,574, 2024.

\bibitem{kirillov2023segment}
A.~Kirillov, E.~Mintun, N.~Ravi, H.~Mao, C.~Rolland, L.~Gustafson, T.~Xiao, S.~Whitehead, A.~C. Berg, W.-Y. Lo \emph{et~al.}, ``Segment anything,'' in \emph{Proceedings of the IEEE/CVF international conference on computer vision}, 2023, pp. 4015--4026.

\bibitem{liu2024grounding}
S.~Liu, Z.~Zeng, T.~Ren, F.~Li, H.~Zhang, J.~Yang, Q.~Jiang, C.~Li, J.~Yang, H.~Su \emph{et~al.}, ``Grounding dino: Marrying dino with grounded pre-training for open-set object detection,'' in \emph{European Conference on Computer Vision}.\hskip 1em plus 0.5em minus 0.4em\relax Springer, 2024, pp. 38--55.

\bibitem{li2021fabric}
C.~Li, J.~Li, Y.~Li, L.~He, X.~Fu, and J.~Chen, ``Fabric defect detection in textile manufacturing: a survey of the state of the art,'' \emph{Security and Communication Networks}, vol. 2021, no.~1, p. 9948808, 2021.

\bibitem{bergmann2021mvtec}
P.~Bergmann, K.~Batzner, M.~Fauser, D.~Sattlegger, and C.~Steger, ``The mvtec anomaly detection dataset: a comprehensive real-world dataset for unsupervised anomaly detection,'' \emph{International Journal of Computer Vision}, vol. 129, no.~4, pp. 1038--1059, 2021.

\end{thebibliography}
\bibliographystyle{IEEEtran}

\end{document}